\pdfoutput=1

\documentclass[11pt]{article}

\usepackage{acl}

\usepackage{times}
\usepackage{latexsym}
\usepackage{graphicx}
\usepackage{subfig}
\usepackage{multirow}
\usepackage{array}
\usepackage{booktabs}
\usepackage{xcolor}
\usepackage{amsmath,amssymb}
\usepackage[geometry]{ifsym}
\usepackage{color}
\usepackage{tikz}
\usetikzlibrary{shapes}

\usepackage{emoji}
\newcommand{\msr}{\emoji[twitter]{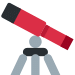}}
\newcommand{\oxford}{\emoji[twitter]{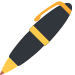}}

\newcolumntype{P}[1]{>{\centering\arraybackslash}p{#1}}

\usepackage[T1]{fontenc}

\usepackage[utf8]{inputenc}

\usepackage{microtype}

\newcommand{\markergreencircle}{\raisebox{0.5pt}{\tikz{\node[draw,scale=0.4,circle,fill=green](){};}}}

\newcommand{\markerredsquare}{\raisebox{0.5pt}{\tikz{\node[draw,scale=0.4,regular polygon, regular polygon sides=4,fill=red](){};}}}

\newcommand{\markerbluetriangle}{\raisebox{0.5pt}{\tikz{\node[draw,scale=0.3,regular polygon, regular polygon sides=3,fill=blue,rotate=0](){};}}}

\title{Revisiting the Compositional Generalization Abilities of \\ Neural Sequence Models}

\author{Arkil Patel \raise1.2ex\hbox{\footnotesize\msr} \quad Satwik Bhattamishra \raise1.2ex\hbox{\footnotesize\oxford} \quad Phil Blunsom \raise1.2ex\hbox{\footnotesize\oxford} \quad Navin Goyal \raise1.2ex\hbox{\footnotesize\msr} \\
	\raise1.2ex\hbox{\footnotesize\msr} Microsoft Research India \\
	\raise1.2ex\hbox{\footnotesize\oxford} University of Oxford \\
	{\tt arkil.patel@gmail.com, navingo@microsoft.com} \\
	{\tt \{satwik.bmishra,phil.blunsom\}@cs.ox.ac.uk} \\
}

\begin{document}
\maketitle
\begin{abstract}
Compositional generalization is a fundamental trait in humans, allowing us to effortlessly combine known phrases to form novel sentences. Recent works have claimed that standard seq-to-seq models severely lack the ability to compositionally generalize. In this paper, we focus on one-shot primitive generalization as introduced by the popular SCAN benchmark. We demonstrate that modifying the training distribution in simple and intuitive ways enables standard seq-to-seq models to achieve near-perfect generalization performance, thereby showing that their compositional generalization abilities were previously underestimated. We perform detailed empirical analysis of this phenomenon. Our results indicate that the generalization performance of models is highly sensitive to the characteristics of the training data which should be carefully considered while designing such benchmarks in future.

 
\end{abstract}

\section{Introduction}

According to the \textit{principle of compositionality}, the meaning of a complex expression (e.g., a sentence) is determined by the meaning of its individual constituents and how they are combined. Humans can effectively recombine known parts to form new sentences that they have never encountered before. Despite the unprecedented achievements of standard seq-to-seq networks such as LSTMs and Transformers in NLP tasks, previous work has suggested that they are severely limited in their ability to generalize compositionally \cite{scan, cp_in_sp}.

\begin{figure}[t]
	\centering
	\includegraphics[scale=0.7, trim=35 30 33 30, clip]{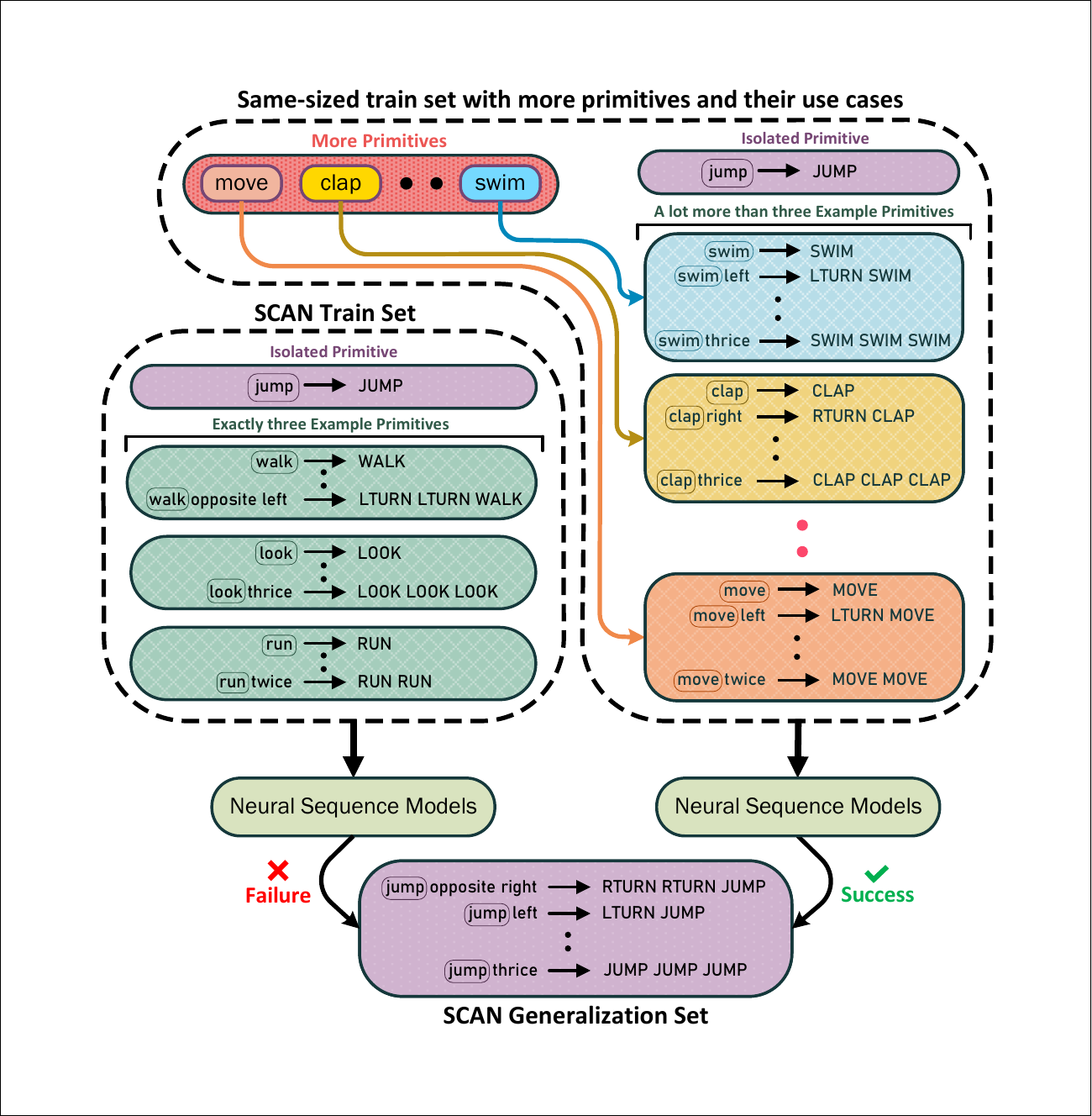}
	\caption{\label{fig:overview} Overview of the SCAN generalization task (left) and our approach (right) that enables standard neural sequence models to generalize compositionally.}
\end{figure}

\textbf{Problem Statement.} Our work relates to a central challenge posed by compositional generalization datasets such as SCAN \cite{scan} and Colors \cite{colors}, which we refer to as \emph{one-shot primitive generalization}: 
The dataset consists of \emph{input-output sentence} pairs (e.g. `walk twice $\rightarrow$ WALK WALK'); input sentences are formed from primitive words (`walk') and function words (`twice') and are generated by a context-free grammar (CFG); output sentences are obtained by applying an interpretation function. Crucially, there is a systematic difference between the train and test splits\footnote{We use the term \emph{systematicity} in the rest of the paper to refer to this difference.}: While the former has a \emph{single} example of an \emph{isolated primitive} (e.g., the primitive definition `jump $\rightarrow$ JUMP' in SCAN), the latter consists of compositional sentences with this isolated primitive (e.g. `jump twice $\rightarrow$ JUMP JUMP'). See Fig.  \ref{fig:overview} (left) for an overview of the task. 

A model with the right inductive bias should generalize on the test data after having seen compositional expressions with other primitives during training. The need for such inductive bias is justified via psychological experiments \cite{colors} indicating that humans do have the ability to generalize on such tasks. Previous works have suggested that seq-to-seq models lack the appropriate inductive bias necessary to generalize on this task since they achieve near-zero accuracies on both SCAN and Colors benchmarks. This has led to the development of many specialized architectures \cite{prim_subs, permutation, ness, lexicon}, learning procedures \cite{lake_meta, titov_meta} and data augmentation methods \cite{geca, kim-rush-aug} to solve the task.

\textbf{Contributions. }The primary claim of our paper is that, contrary to prior belief, neural sequence models such as Transformers and RNNs do have an inductive bias\footnote{However, note that this inductive bias is not as strong as that of specialized architectures designed for these tasks.} to generalize compositionally which can be enabled using the right supervision. \textbf{(i)}   We show that by making simple and intuitive changes to the training data distribution, standard seq-to-seq models can achieve high generalization performance even with a training set of size less than 20\% of the original training set. In particular, if we incorporated examples with more novel primitives in the training set without necessarily increasing the size of the training set (see right part of Fig. \ref{fig:overview}), then the generalization performance of standard seq-to-seq models improves and reaches near-perfect score after a certain point. Our results also exemplify the importance of the training distribution apart from architectural changes and demonstrate that providing the right supervision can significantly improve the generalization abilities of the models.
\textbf{(ii)} We investigate the potential cause behind the improvement in generalization performance and observe that the embedding of the isolated primitive becomes more similar to other primitives when the training set has higher number of primitives and their use cases.
\textbf{(iii)} To understand the phenomenon better, we characterize the effect of different training distributions and model capacities. Our results show that the parameters of the experimental setting play a crucial role while evaluating the generalization abilities of models.



%

\section{Enabling Generalization by Providing the Right Supervision}\label{sec:prim}


\textbf{Setup.} We focus on the SCAN and Colors datasets.\footnote{Results on COGS \cite{cogs} can be found in Appendix \ref{sec:cogs}.} Both these datasets have exactly one \emph{isolated primitive}. We refer to all other primitives (i.e., those that are also composed with other words to form sentences in the training set) as \emph{example primitives}. Both the SCAN and Colors training sets have exactly three example primitives. The training set of SCAN has $13.2$k examples while the test set has $7.7$k examples. Colors has just $14$ training examples and $8$ test examples. More details on implementation and datasets can be found in Appendix \ref{sec:implementation} \& \ref{sec:datasets}. Our source code is available at \href{https://github.com/arkilpatel/Compositional-Generalization-Seq2Seq}{ https://github.com/arkilpatel/Compositional-Generalization-Seq2Seq}.


\textbf{Adding More Primitives. } We modify the training set such that the number of distinct example primitives present in the dataset is higher. To do so, we add new primitives to the language which are simply random words (e.g., `swim', `clap', etc.) that have the same semantics and follow the same grammar rules as other existing primitives (see Fig. \ref{fig:overview} (right) for illustration). These new primitives act as example primitives in our training set. For SCAN, we control the size of the training set such that it is at most the size of the original dataset.\footnote{The training set size $|T|$ is kept fixed by discarding original examples and adding $(|T|/\#primitives)$ examples per primitive. Because of extremely small data size, we cannot do this for Colors while also trying to illustrate our idea.} To generate the training set, we randomly sample the examples from the new grammar and discard all compositional sentences with the isolated primitive. For each example primitive and the isolated primitive, a primitive definition (such as `walk $\rightarrow$ WALK') is also added to the training set. The test set is untouched and remains the same.


\begin{figure}[t]
	\centering
	\includegraphics[scale=0.55, trim=8 5 10 5, clip]{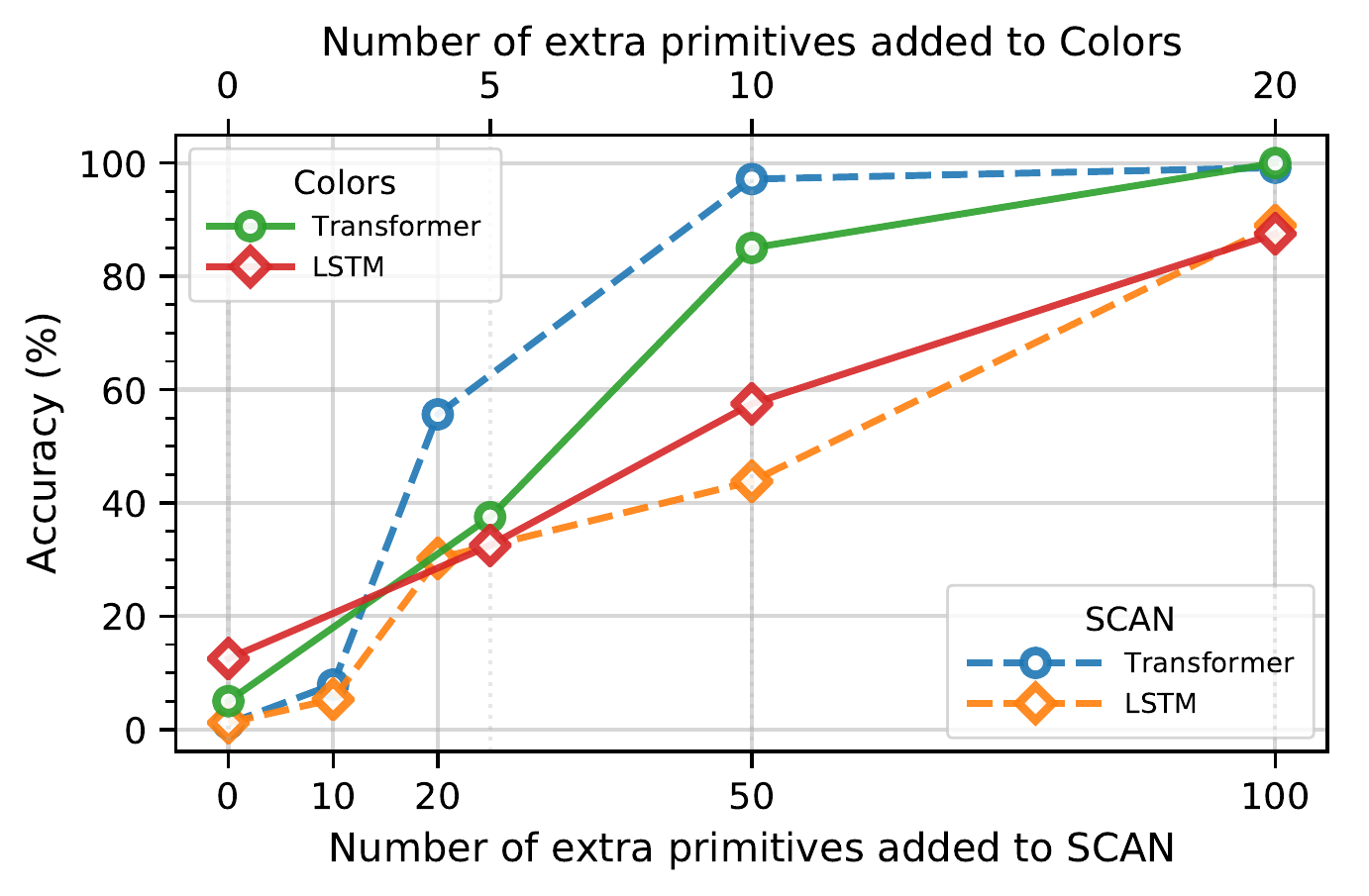}
	\caption{\label{fig:more_primitives} Generalization performance ($\uparrow$) on SCAN and Colors improves with higher number of example primitives in the training set.}
\end{figure}


\textbf{Main Observation.} Fig. \ref{fig:more_primitives} shows the generalization performance of Transformer and LSTM based seq-to-seq models. We observe that there is a clear trend of improvement in compositional generalization as we increase the number of example primitives and their use cases. It is surprising to see that on SCAN, Transformers perform on par with some recently proposed specialized architectures \cite{prim_subs, permutation} and even better than certain architectures \cite{syntactic}.


\begin{figure}[t]
	\centering
	\includegraphics[scale=0.48, trim=8 8 10 5, clip]{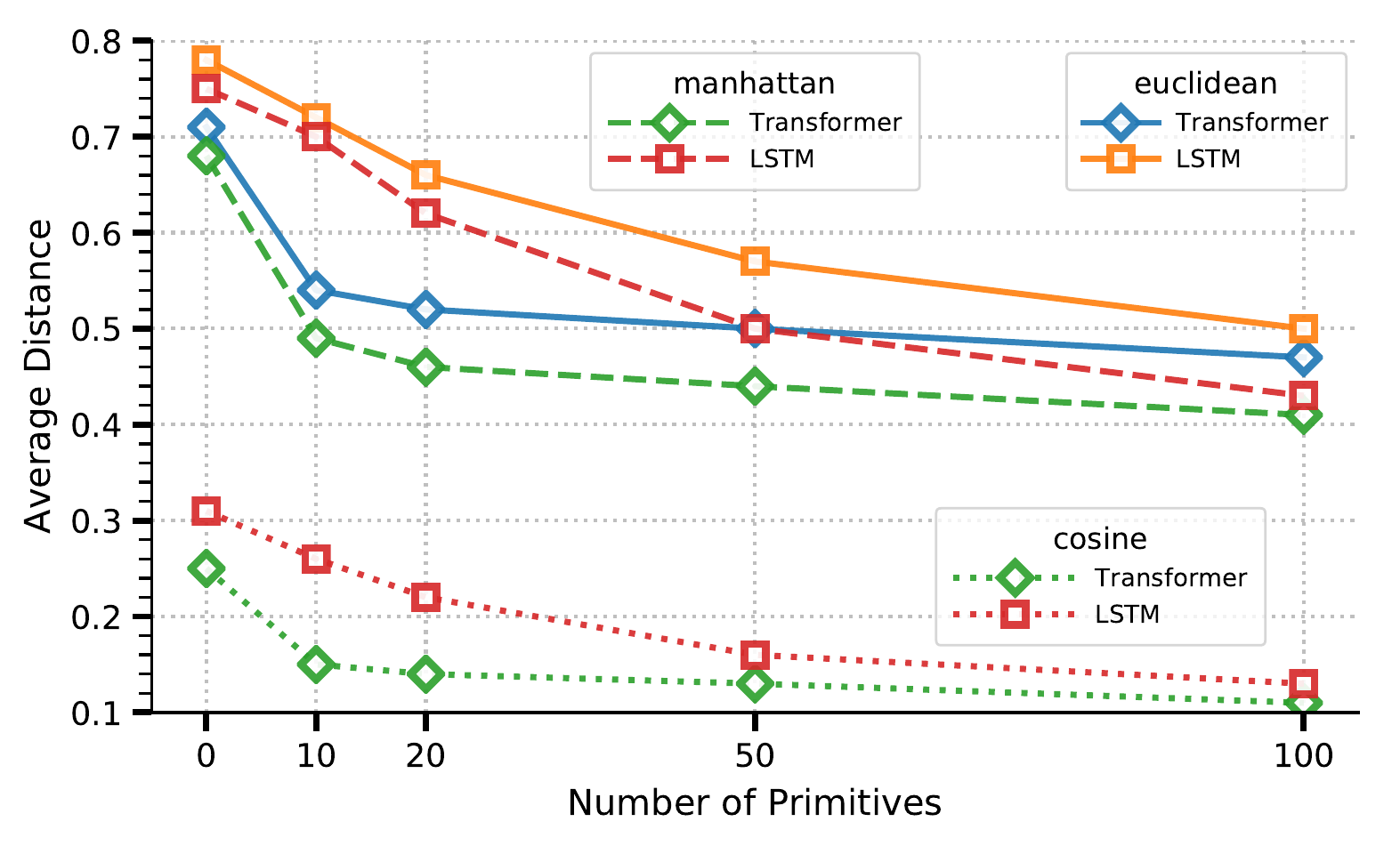}
	\caption{\label{fig:scan_similarity} Measuring the distance of embedding of \emph{isolated primitive} with embeddings of example primitives for learned Transformer and LSTM models as we increase the number of example primitives in SCAN.}
\end{figure}


\textbf{Implication.} Since the training set still contains only one non-compositional example with the isolated primitive\footnote{Note that our results also hold when there are multiple isolated primitives in the dataset at the same time. This is discussed in Appendix \ref{sec:multiple_iso_prims}.} and the test set is untouched, one-shot primitive generalization setting is preserved. Hence our results clearly show that standard neural sequence models have `some' inductive bias required to generalize on such out-of-distribution tasks even if it is not as strong as that of specialized architectures designed primarily to solve these tasks. Our results are in contradiction to previously suggested limitations of standard seq-to-seq models in terms of primitive generalization \cite{scan,cp_in_sp,baroni_opinion}. While it is important to develop architectures with better compositional generalization abilities, we wish to highlight that synthetic benchmarks such as SCAN require a model with very strong inductive biases and tend to underestimate the generalization abilities of baseline models. 


While we have shown that these models can generalize from one-shot exposure to primitive definitions, our results also hold for the more general case where the one-shot exposure of the primitive is in a sentence (e.g. `jump twice $\to$ JUMP JUMP'). More details regarding these experiments can be found in Appendix \ref{sec:implicit}.

\textbf{Prior Work.} Note that our work is unrelated to previous works that propose data augmentation approaches for compositional generalization tasks \cite{geca, kim-rush-aug, andreas-RandR}. (1) The datasets created by some of these augmentation methods do not preserve the systematic differences between train and test sets, while our datasets do.\footnote{We discuss this in more detail in Appendix \ref{sec:sim_work}.} (2) The objective of these works was to devise a method to improve compositional generalization performance whereas the focus of our work is not to develop a general method; rather we want show that \textbf{baseline seq-to-seq models are capable of generalizing compositionally even without breaking systematicity}. (3) These methods add additional data resulting in datasets of larger sizes whereas we control for data size.

\subsection{Analyzing the Embedding of the Isolated Primitive}

\begin{figure}[t]%
	\centering
	\subfloat[\centering No extra primitives]{{\includegraphics[scale = 0.32, trim=5 10 20 10, clip]{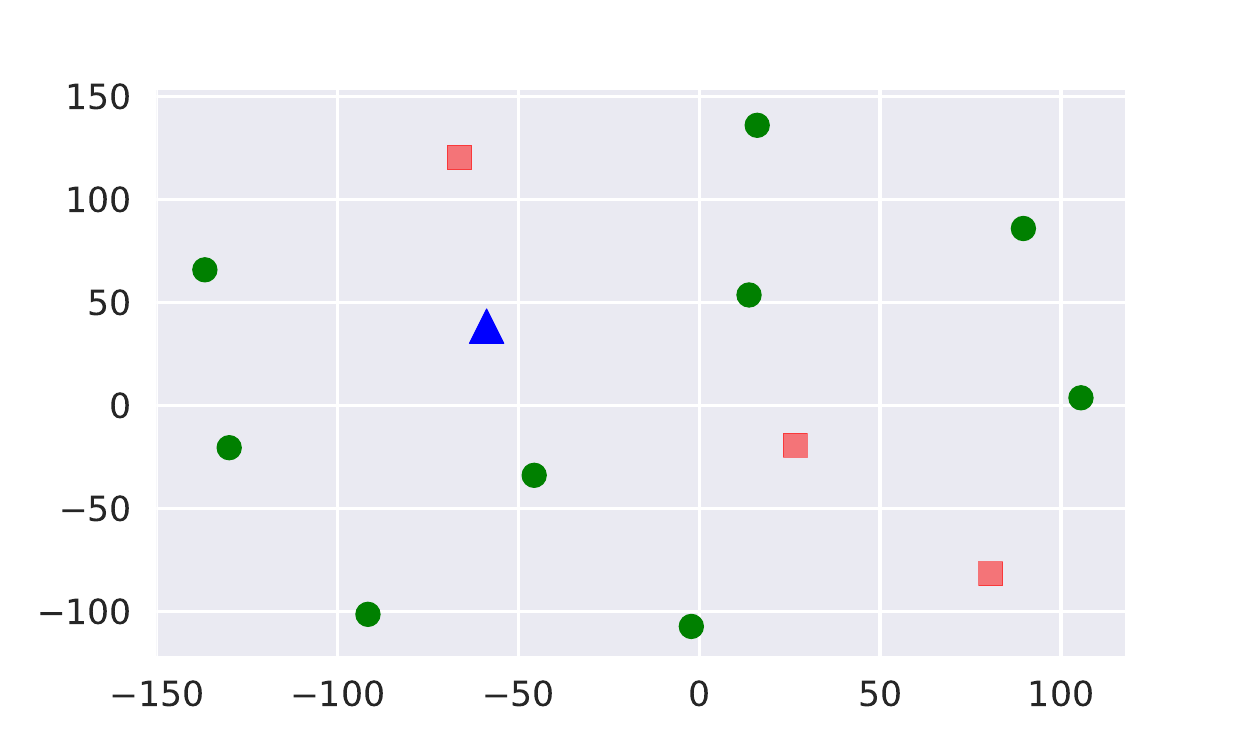} }}%
	\subfloat[\centering 10 extra primitives]{{\includegraphics[scale = 0.32, trim=5 10 20 10, clip]{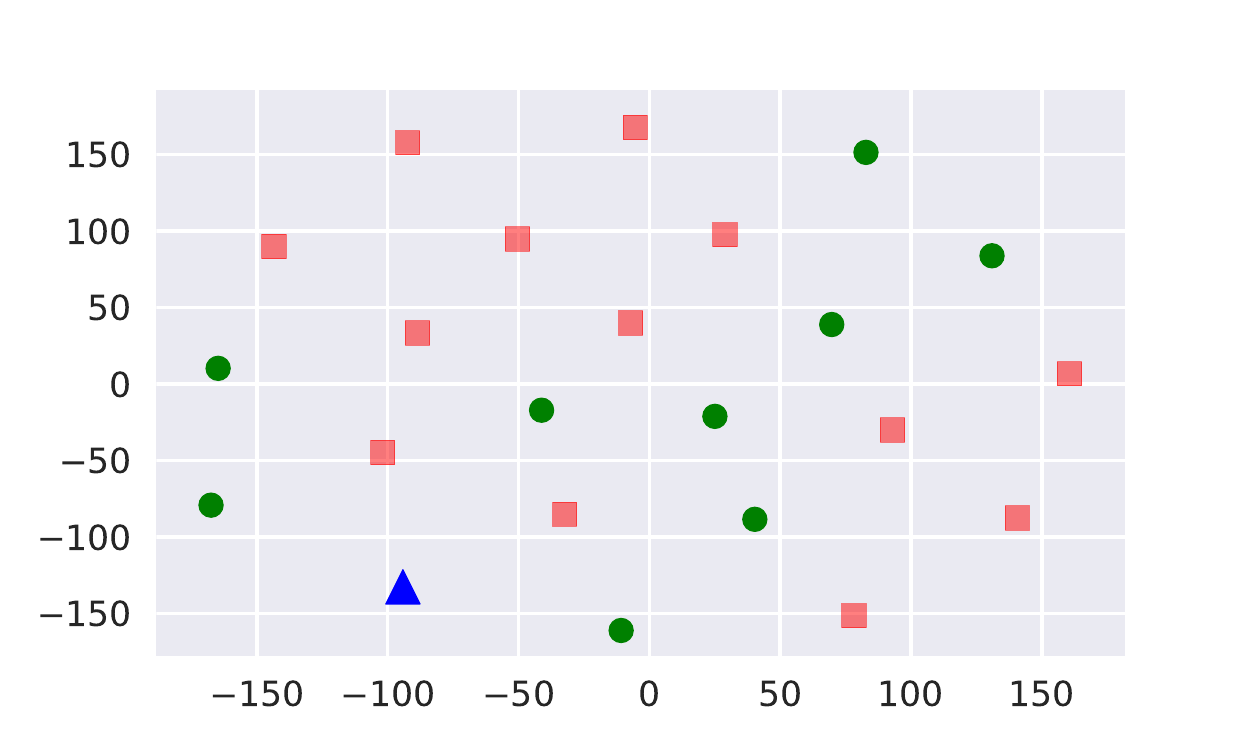} }}%
	\qquad
	\subfloat[\centering 20 extra primitives]{{\includegraphics[scale = 0.32, trim=5 10 20 10, clip]{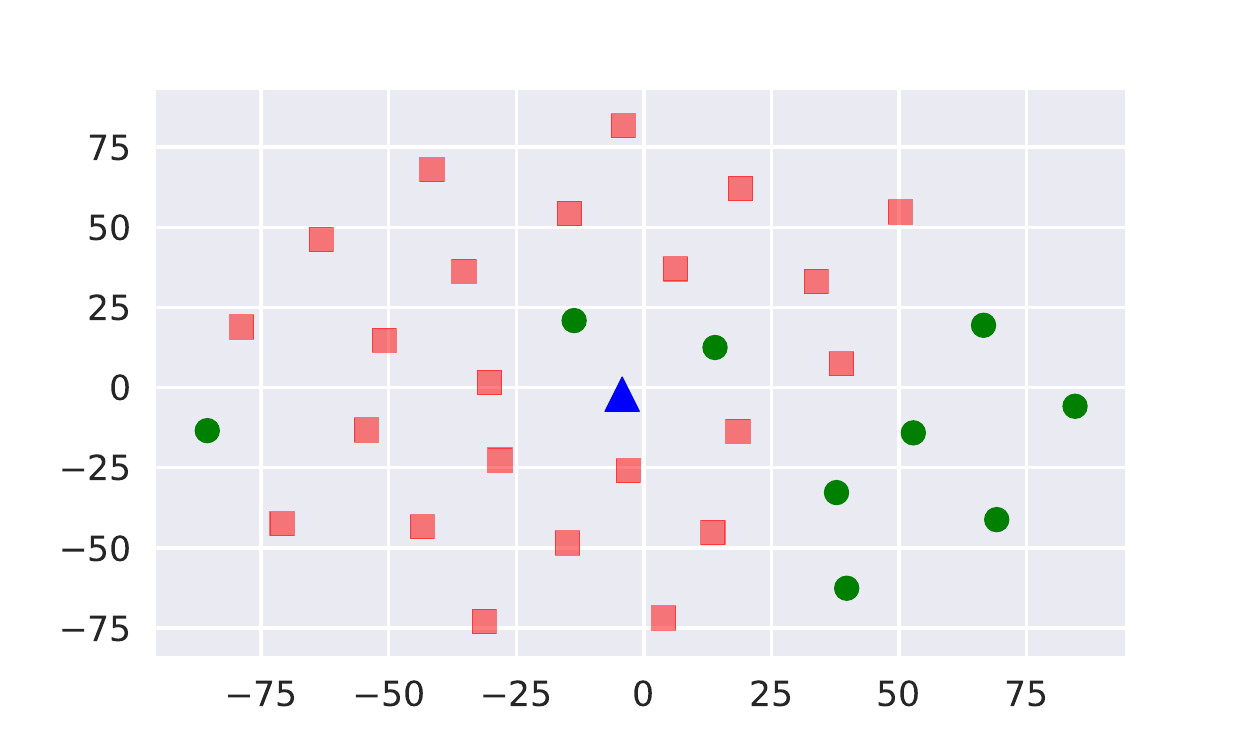} }}%
	\subfloat[\centering 50 extra primitives]{{\includegraphics[scale = 0.32, trim=5 10 20 10, clip]{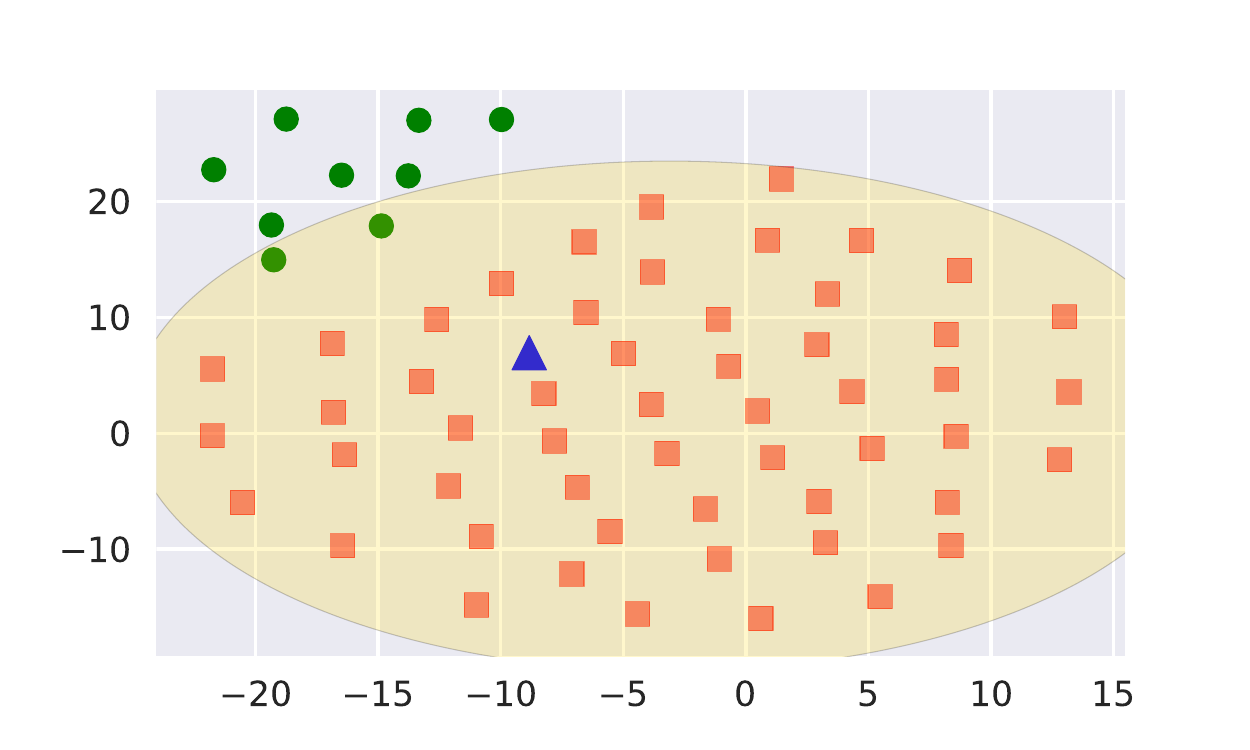} }}%
	\caption{Visualizing the $t$-SNE reduced embeddings of isolated primitive (\markerbluetriangle), example primitives (\markerredsquare) and non-primitives (\markergreencircle) from a learned Transformer model as we increase number of example primitives in SCAN.}%
	\label{fig:scan_transformer_tsne}%
\end{figure}

Our results raise the question: Why do Transformers and LSTMs generalize better when the training data has more example primitives? Compositional generalization in our setting requires a model to learn to apply the same rules to the isolated primitive as it does to the other example primitives. Thus, we analyze the change in the learned embedding of the isolated primitive (such as `jump') with respect to other primitives in different settings. 

In particular, we compare the average distance with other primitives before and after adding certain number of primitives to training data (this is the same setting that was explained earlier in this section). We find that as we increase the number of example primitives in the training set, the embedding of the isolated primitive gets closer to the example primitives (Fig. \ref{fig:scan_similarity}) in terms of Euclidean, Manhattan and Cosine distances. If the embedding of the isolated primitive is closer to the embeddings of the other primitives, then the model is more likely to operate over it in a similar fashion and apply the same rules as it does over the other primitives. 

This phenomenon is also illustrated in $t$-SNE plots (Fig. \ref{fig:scan_transformer_tsne}) of the learned embeddings where the embedding of the isolated primitive seems closer to the embeddings of the example primitives when there are more example primitives in the dataset. Hence, a possible reason behind improved generalization performance could be the difference in the learned embeddings.\footnote{More fundamental reasons for difference in learned embeddings, such as learning dynamics, are beyond our scope.} Additional results with the LSTM model and Colors dataset can be found in Appendix \ref{sec:embedding}.

%
%
%
%
%

\section{Exploring the Impact of the Parameters of the Experimental Setup}\label{sec:robustness}

\subsection{Impact of Training Distributions}\label{sec:train_dist}

\begin{figure}[t]%
	\centering
	\subfloat[\centering Other Distributions]{{\includegraphics[scale = 0.55, trim=5 8 15 10, clip]{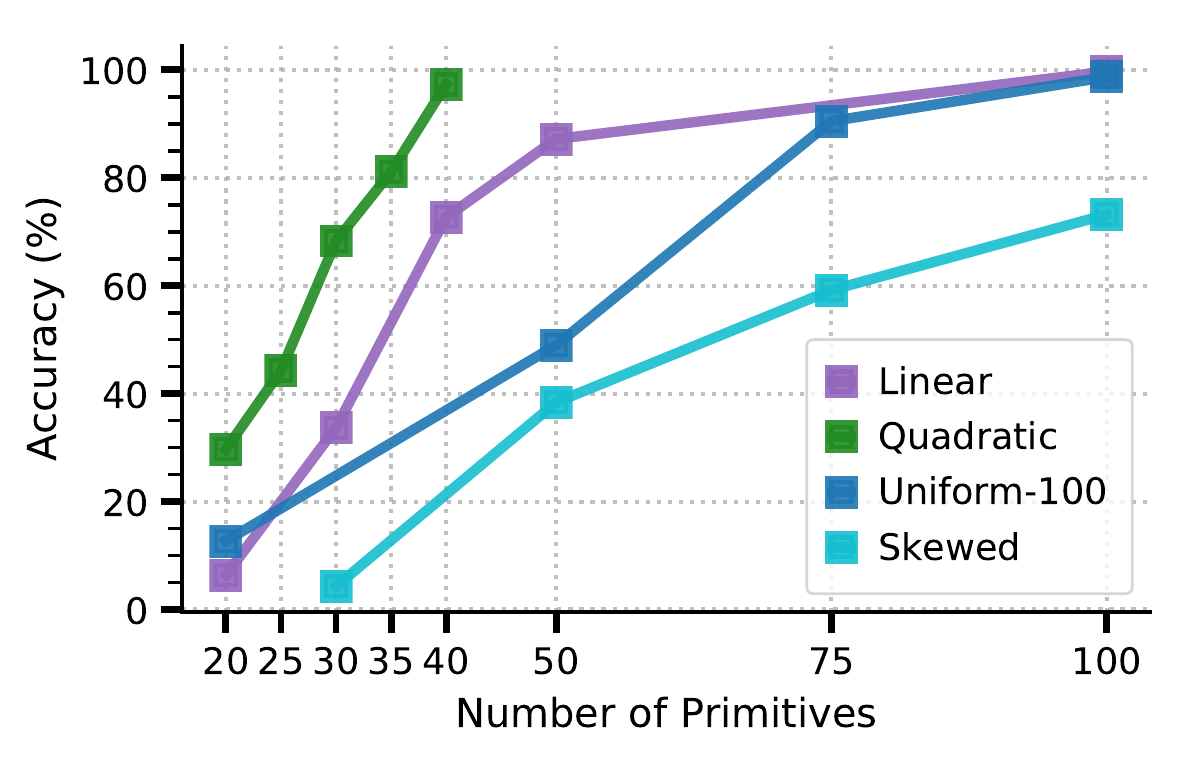} \label{fig:scan_dist} }}
	
	\subfloat[\centering Uniform Distribution]{{\includegraphics[scale = 0.6, trim=0 0 5 5, clip]{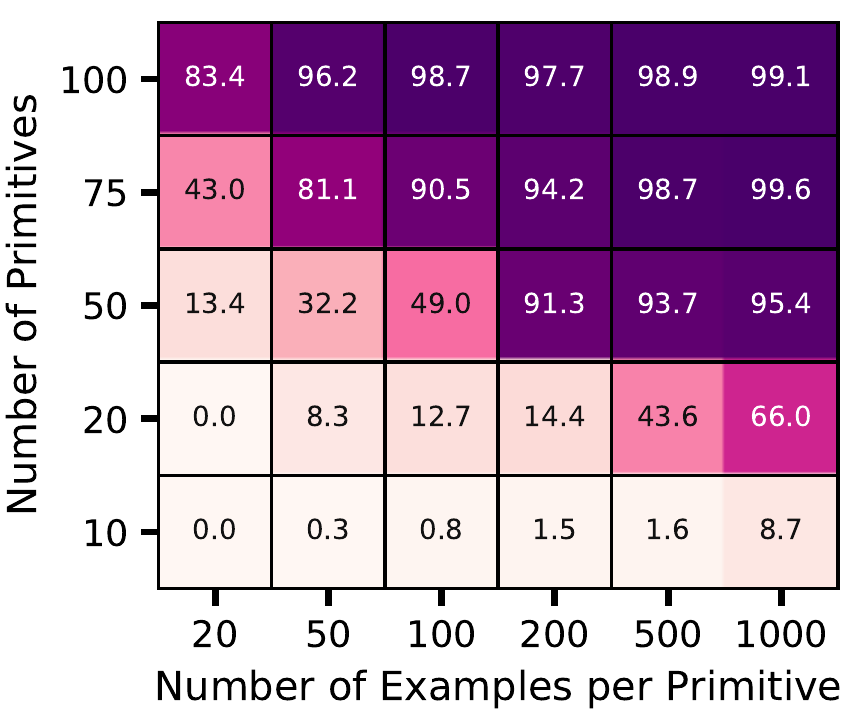} \label{fig:scan_grid_uni} }}
	
	\caption{Measuring the generalization performance of Transformer on different types of training set distributions of the SCAN dataset.}%
	\label{fig:scan_transformer_distribution}%
\end{figure}


 In this section, we analyze the influence of different training distributions on the generalization performance of the model. In the previous experiments, the data generating distribution was uniform over all possible samples. Here, we alter the training data distribution by varying the number of examples for each example primitive. The test set remains unchanged and there will still be only one non-compositional example of the isolated primitive (i.e., the primitive definition) in the training set. We experiment with linearly, quadratically and exponentially increasing probability distribution functions. For instance, in the quadratically increasing case, a training set with $10$ example primitives will have one example primitive with $1$ compositional example, the next one with $4$ compositional examples, another one with $9$ compositional examples and so on.\footnote{In all experimental setups considered in this paper, each example primitive will always have a primitive definition in the training set.} Similarly, in the exponentially increasing case (which we also call `skewed'), 10\% example primitives have 500 compositional examples each, 30\% have 10 compositional examples each and the remaining have just one compositional example each in the training set. The general idea is that all the example primitives do not have equal representation in the training data. Upon training the models on different distributions, we observed that the models generalize well even with fewer number of example primitives when their distribution is linearly or quadratically increasing (Fig. \ref{fig:scan_dist}). On the other hand models struggle to generalize when the distribution is skewed. In that case, most primitives appear in only one or very few compositional sentences in the training data. The failure to generalize on such data implies that extra primitives must be added as part of multiple compositional sentences; just adding the primitive definition or a single example for each example primitive does not help the model to leverage it.


We then try to characterize the relationship between the number of example primitives and the amount of data required for the model to generalize well on the test data, when the example primitives are uniformly distributed. We create different training sets by varying the total number of example primitives, $\#primitives$; for each example primitive, we draw $\#examples$ number of samples uniformly from the CFG. Fig. \ref{fig:scan_grid_uni} shows the generalization performance of Transformers for each of these training sets. The size of each training set is the product of the row and column values ($\#primitives \times \#examples $). As expected, the upper-right triangle has higher scores indicating that the sample requirement decreases as we add more primitives to the dataset. Surprisingly, the top-left cell indicates that Transformers can achieve high performance even with $2$k training examples which is less than \textbf{20\%} of the original SCAN training set. Additional results with the LSTM model can be found in Appendix \ref{sec:train_dist_appendix}.

\subsubsection{Understanding Transferability}

We wish to check whether the inductive bias that is enabled when a model is trained on more number of example primitives can be transferred to a scenario where the number of example primitives is limited. We create a \emph{pretraining} set with 50 example primitives uniformly distributed, each of them having 200 examples. The \emph{finetuning} set is the original SCAN training set and the test set is the original SCAN test set. The model is first trained from scratch on the pretraining set and then finetuned on the finetuning set. 

We find that if we allow all the parameters of the Transformer model to be updated during the finetuning phase on the original SCAN training set, then the model generalizes very poorly. On the other hand, when we freeze the weights of the encoder and decoder after the pretraining phase, and only allow the embedding and output layers to be updated, then the model generalizes near-perfectly on the test set. Our hypothesis is that in the latter setting, the task becomes simpler for the model since it only has to align the embeddings of the newly seen primitives in the finetuning phase with the embeddings of the primitives seen during the pretraining phase. This experiment also indicates that the previously learned rules during pretraining can help a model to compositionally generalize on novel primitives.


 



\begin{figure}[t]
	\centering
	\includegraphics[scale=0.55, trim=8 8 10 5, clip]{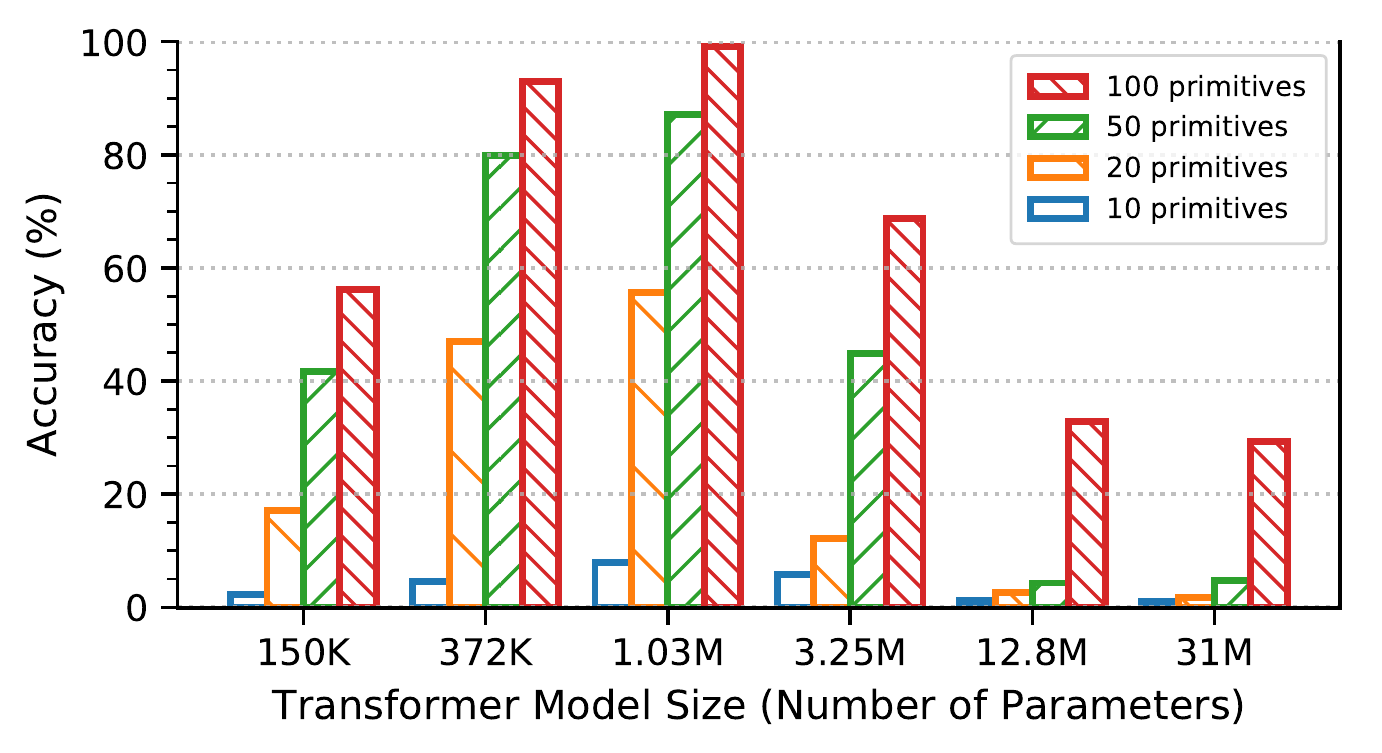}
	\caption{\label{fig:scan_trans_capacity} Measuring the generalization performance of a Transformer of varying capacity across increasing number of primitives in the SCAN training set.}
\end{figure}

\subsection{Impact of Model Capacity}\label{sec:capacity}

We analyze the relationship between the model capacity and the number of example primitives in the training set. We vary the number of primitives as per the description in Section \ref{sec:prim}. We evaluate the generalization performance of the models while gradually increasing the number of parameters by increasing the size of its embeddings and intermediate representations. For each experiment, we exhaustively finetune the rest of the hyperparameters (e.g., dropout, learning rate, batch size, etc.) to select the best model. Looking at Fig. \ref{fig:scan_trans_capacity}, we observe a general trend in which the model starts to overfit and has poor generalization performance as we increase the model size. Note that all these model configurations are able to achieve near-perfect accuracies on the SCAN random split that does not test for compositional generalization. This shows that carefully controlling the model size is important for achieving compositional generalization. On such small datasets, larger models might simply memorize the input-output mappings in the training set. Indeed, such memorization has been cited as a potential reason to explain why models fail at compositional generalization \cite{titov_meta}. We also find that as we increase the number of example primitives, the models are less susceptible to overfitting and achieve relatively better generalization performance. Additional results with the LSTM model and Colors dataset can be found in Appendix \ref{sec:capacity_appendix}.

\section{Conclusion}

While it is essential to make progress in building architectures with better compositional generalization abilities, we showed that the generalization performance of standard seq-to-seq models (often used as baselines) is underestimated. A broader implication of our experiments is that although systematicity must be preserved when designing such benchmarks, it is imperative to carefully explore different parameters associated with the experimental setup to draw robust conclusions about a model's generalization abilities.

%


\section*{Acknowledgements}

We thank the anonymous reviewers for their constructive comments. We would also like to thank Kabir Ahuja, Zihuiwen Ye and our colleagues at Microsoft Research India for their valuable feedback and helpful discussions.

\bibliography{citations}
\bibliographystyle{acl_natbib}

\clearpage
\newpage
\appendix

\section{Implementation Details}\label{sec:implementation}

\begin{table*}[t]
	\scriptsize{\centering
		\begin{tabular}{p{7.5em}P{7em}P{7em}P{7em}P{7em}P{7em}}
			\toprule
			&\multicolumn{2}{c}{\textbf{SCAN}} &\multicolumn{2}{c}{\textbf{COLORS}}
			&\textbf{COGS} \\ 
			\cmidrule(lr){2-3}\cmidrule(lr){4-5}\cmidrule(lr){6-6}
			\textbf{Hyperparameters} & Transformer & LSTM & Transformer & LSTM & Transformer \\
			\midrule
			Embedding Size & [64, \textbf{128}, 256] & [64, \textbf{128}, 256] & [16, \textbf{32}, 64] & [16, \textbf{32}, 64] & [\textbf{384}, 512] \\
			
			Hidden/FFN Size & [\textbf{256}, 512] & [\textbf{64}, 128] & [16, \textbf{32}, 64] & [16, 32, \textbf{64}] & [\textbf{512}, 1024] \\
			
			Heads & [\textbf{2}, 4] & N/A & [\textbf{4}, 8] & N/A & [2, \textbf{4}] \\
			
			Number of Layers & [2, \textbf{3}] & [1, \textbf{2}] & [\textbf{2}, 3] & [1, \textbf{2}] & [\textbf{2}, 3] \\
			
			Learning Rate & [3e-4, \textbf{5e-4}, 8e-4] & [5e-3, \textbf{8e-3}, 1e-2] & [\textbf{8e-4}, 1e-3] & [5e-3, \textbf{8e-3}, 1e-2] & [3e-4, \textbf{5e-4}, 8e-4] \\
			
			Batch Size & [\textbf{128}, 256] & [128, \textbf{256}] & [\textbf{1}, 2] & [\textbf{1}, 2] & [\textbf{128}, 256] \\
			
			Dropout & [\textbf{0.1}, 0.2] & [\textbf{0.1}, 0.2] & [\textbf{0.1}, 0.2] & [\textbf{0.1}, 0.2] & [\textbf{0.1}, 0.2] \\
			
			\midrule
			
			
			Epochs & 150 & 150 & 150 & 150 & 150 \\
			
			Avg Time/Epoch & 30 & 40 & 2 & 3 & 60 \\
			
			\bottomrule
		\end{tabular}
		\caption{\label{tab:hyperparams}Different hyperparameters and the values considered for each of them in the models. The best hyperparameters for each model for all the datasets (with maximum number of primitives of all the settings studied in this paper) are highlighted in bold. Average Time/Epoch is measured in seconds.}
	}
\end{table*}

We use 8 NVIDIA Tesla P100 GPUs each with 16 GB memory to run our experiments. All models are implemented in PyTorch \cite{pytorch}. We do not use any pretrained models and all embeddings are learnt from scratch. Parameters are updated using Adam optimization. All results are an average of 5 different runs with random seeds. The dataset-specific hyperparameters used for each model are shown in Table \ref{tab:hyperparams}.

\section{Primitive Generalization Datasets}\label{sec:datasets}

In this paper, we show results on three datasets that evaluate primitive generalization.

\textbf{SCAN} \cite{scan} is a supervised sequence-to-sequence semantic parsing task wherein the natural language input command has to be transformed to the corresponding set of actions. The complete dataset consists of all the commands (a total of 20,910) generated by a phrase-structure grammar and the corresponding sequence of actions, produced according
to a semantic interpretation function. The benchmark consists of 4 splits: random, add jump, turn left and length. We work on the `add jump' split which was designed to test primitive generalization. In this split, the test set (size: 7706) is made up of all the compositional sentences with the primitive `jump' (which we refer to as the \emph{isolated primitive}). The train set (size: 13,204\footnote{The dataset released by \cite{scan} is of size 14,670 which has many repetitions of the `jump $\rightarrow$ JUMP' primitive definition. In this work, we remove all these repetitions since they do not significantly help in generalization.}) has just one example of the isolated primitive (i.e. the primitive definition `jump $\rightarrow$ JUMP') and other examples demonstrating the definitions and compositions of the three other primitives (which we refer to as the \emph{example primitives}). Table \ref{tab:scan} illustrates the task.

\begin{figure}[t]
	\centering
	\includegraphics[scale=0.23, trim=5 5 5 5, clip]{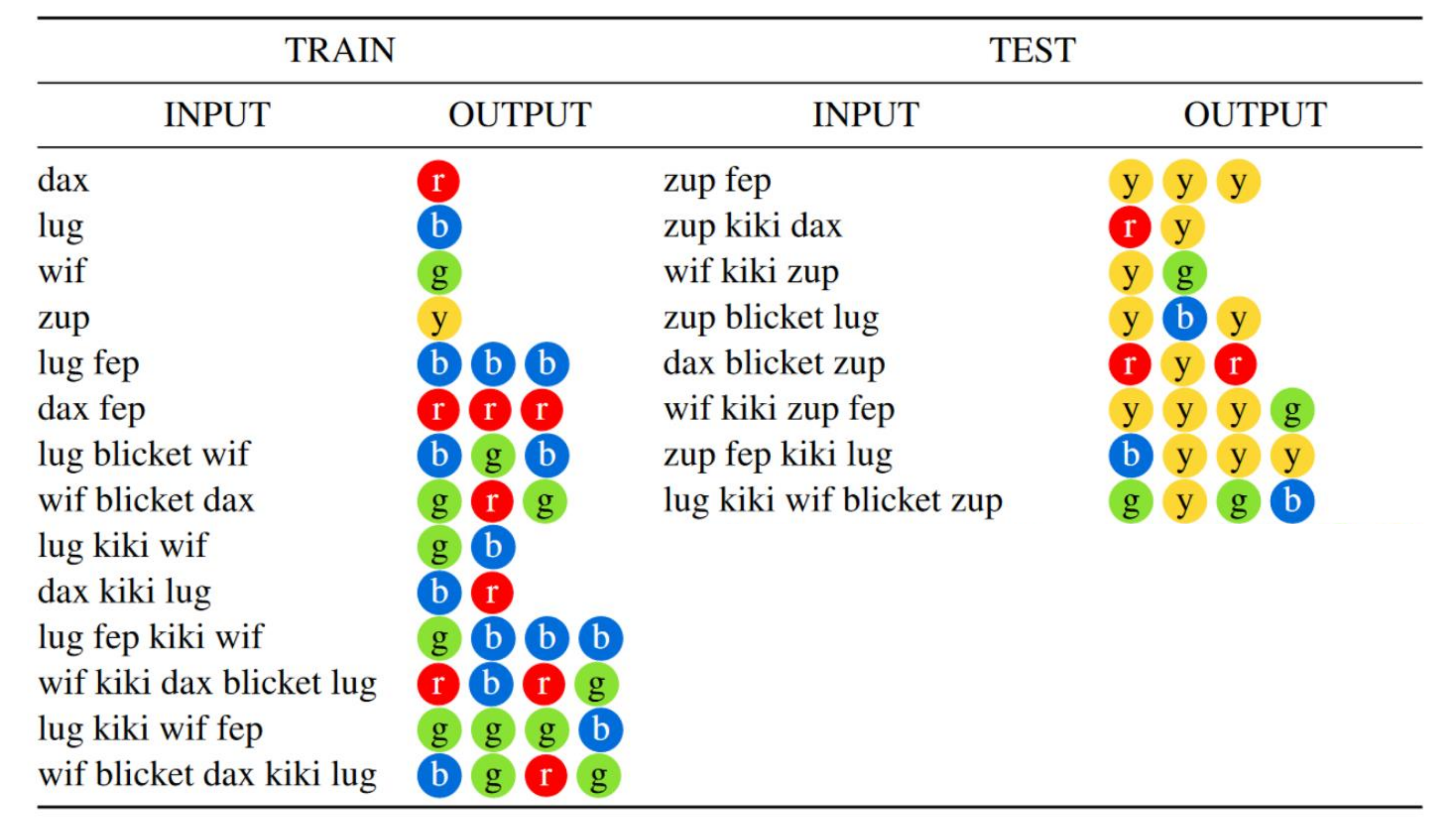}
	\caption{\label{fig:colors} The primitive generalization task in Colors\footnotemark. Note that the test set does not contain the two length generalization examples.}
\end{figure}

\textbf{Colors} \cite{colors} is a sequence-to-sequence task that was designed to
measure human inductive biases. Apart from the challenge of primitive generalization, this dataset poses an additional challenge of low-resource learning for neural sequence models. The train set has just 14 examples that are either primitive definitions of the four primitives or examples with compositions of the three example primitives and three operations (concatenation, repetition and wrapping). The test set has 8 examples\footnote{The original dataset has two additional examples which evaluate length generalization. Since we focus only on primitive generalization, we do not evaluate on these.} with compositions of the isolated primitive (`zup'). Fig. \ref{fig:colors} illustrates the task.

\textbf{COGS} \cite{cogs} is a semantic parsing task of mapping English natural language sentences to their corresponding logical forms. Apart from primitive generalization, COGS also evaluates other types of systematic generalization such generalizing to higher depths or generalizing to novel syntactic structures. The size of the train set is 24,155 and that of the test set is 21,000.

\begin{table}[t]
	\footnotesize{\centering
		\begin{tabular}{m{8.5em}m{12.5em}}
			\toprule
			\small{\textsc{Train:} } & \\
			\midrule
			\textbf{jump} & JUMP \\
			run after run left & LTURN RUN RUN \\
			run & RUN \\
			look left twice & LTURN LOOK LTURN LOOK \\
			\midrule
			\small{\textsc{Test:} } & \\
			\midrule
			\textbf{jump} twice after look & LOOK JUMP JUMP \\
			turn left and \textbf{jump} & LTURN JUMP \\
			\textbf{jump} right twice & RTURN JUMP RTURN JUMP \\
			\bottomrule
		\end{tabular}
		\caption{\label{tab:scan} An illustration of the primitive generalization task in SCAN.}
	}
\end{table}

\footnotetext{Image taken from \citet{lexicon}.}

\begin{figure}[t]
	\centering
	\includegraphics[scale=0.5, trim=8 8 10 5, clip]{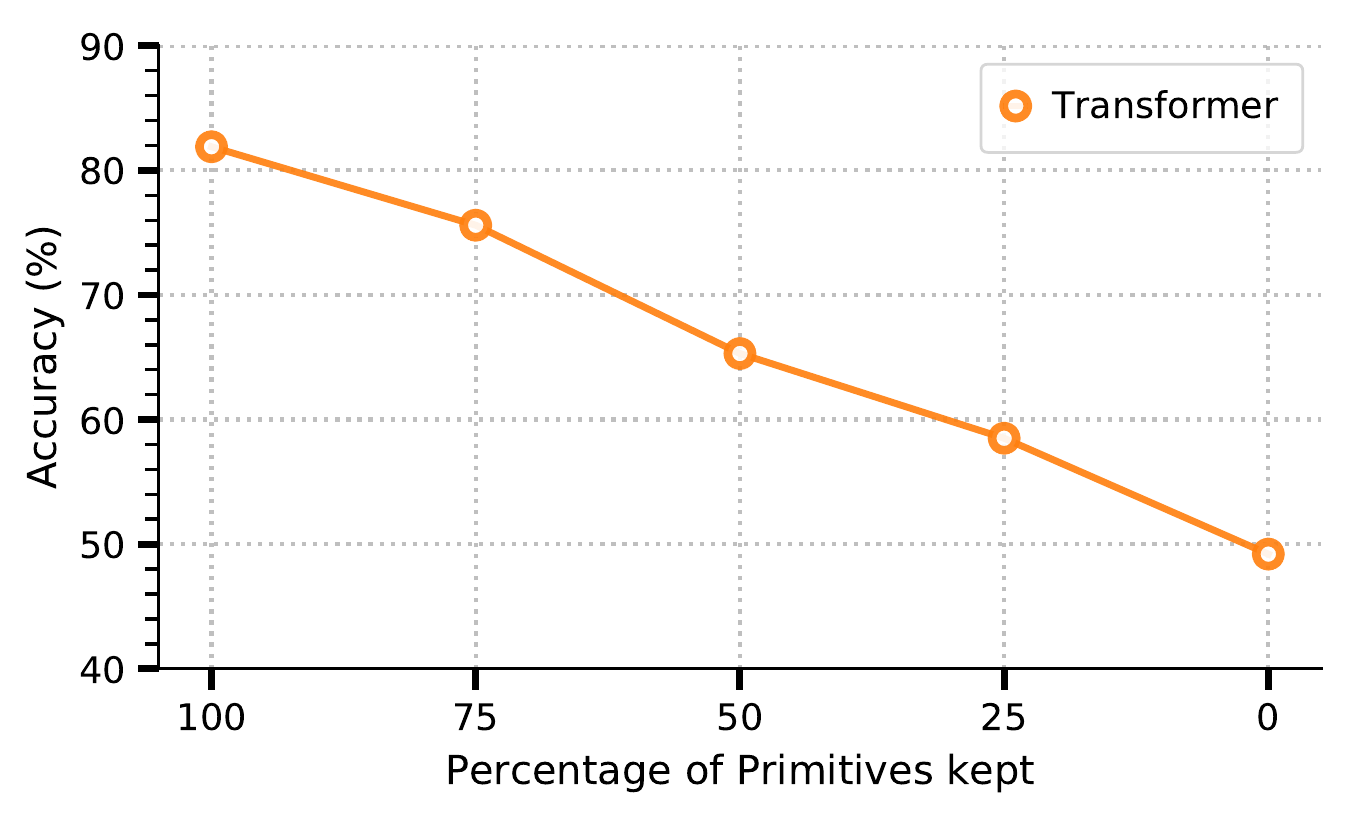}
	\caption{\label{fig:cogs_primitives} Decrease in generalization performance on our COGS primitive generalization test set with a decrease in the percentage of example primitives and their use cases present in the train set.}
\end{figure}

%
%
%
%

\section{Removing Primitives Hurts Generalization on COGS}\label{sec:cogs}

Unlike SCAN and Colors, both of which have a single isolated primitive and only 3 example primitives, COGS has 3 isolated primitives - a verb, a common noun and a proper noun which are supported by 80 verbs, 40 common nouns and 20 proper nouns as example primitives. We hypothesize that this high number of example primitives might be one of the reasons behind the high performance of Transformers on COGS \cite{devil, google_making}, as far as primitive generalization is concerned.

To validate our hypothesis, we systematically reduce the number of example primitives in COGS and evaluate the model. The test set of COGS focusing on primitive generalization consists of 5000 examples. If we directly start removing the primitives from the train set, we risk having out-of-vocabulary tokens in the test set. Hence we select a portion of the test set of size 1218 which exludes 129 example primitives. We will hold this test set fixed and vary the percentage of the 129 example primitives to be inserted in the train set. For each example primitive, samples are drawn uniformly from the original COGS train set. Note that even though the number of example primitives and their use cases will vary in the train set, we control the total train set size to be always 2500 for fair evaluation.

The results of our experiment can be seen in Fig. \ref{fig:cogs_primitives}. We see a clear trend of decrease in generalization performance as we decrease the number of example primitives and their use cases. This is in tandem with the results shown in Section \ref{sec:prim} and further validates the idea that providing more example primitives and their use cases helps neural sequence models generalize on the primitive generalization task. Our results help explain that the gap in performance of neural sequence models on primitive generalization tasks in COGS and primitive generalization tasks in SCAN or Colors is at least partially caused by the difference in the number of example primitives and their use cases in these datasets.

\section{Implicit Word Learning}\label{sec:implicit}

\begin{table}[t]
	\footnotesize{\centering
		\begin{tabular}{P{3.5em}P{18em}}
			\toprule
			\textsc{Complexity} & 			\textsc{Sentence} \\
			\midrule
			1 & \textbf{jump} twice \\
			2 & \textbf{jump} thrice and look \\
			3 & run twice after \textbf{jump} opposite left \\
			4 & \textbf{jump} around left and walk opposite left twice \\
			\bottomrule
		\end{tabular}
		\caption{\label{tab:scan_complexity} Sentences of varying complexities featuring the isolated primitive `jump'.}
	}
\end{table}

\begin{figure*}[t]%
	\centering
	\subfloat[\centering No extra primitives]{{\includegraphics[scale = 0.4, trim=5 10 20 10, clip]{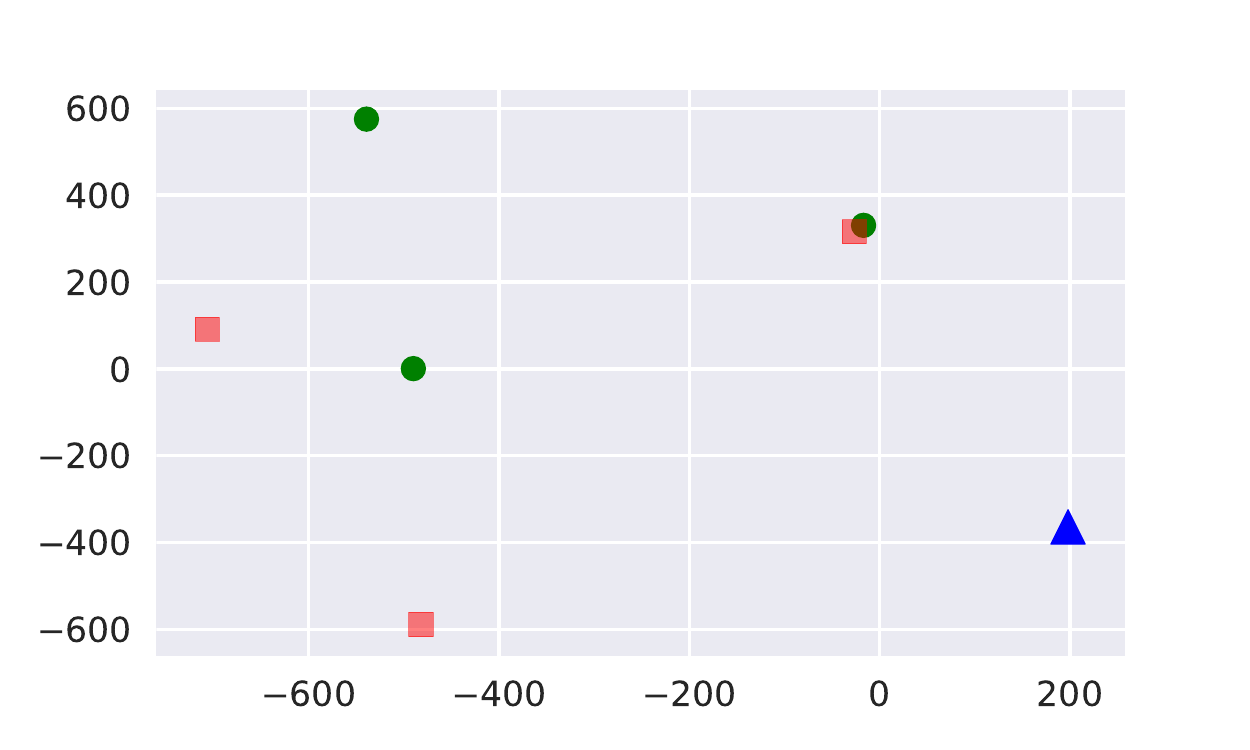} }}%
	\subfloat[\centering 10 extra primitives]{{\includegraphics[scale = 0.4, trim=5 10 20 10, clip]{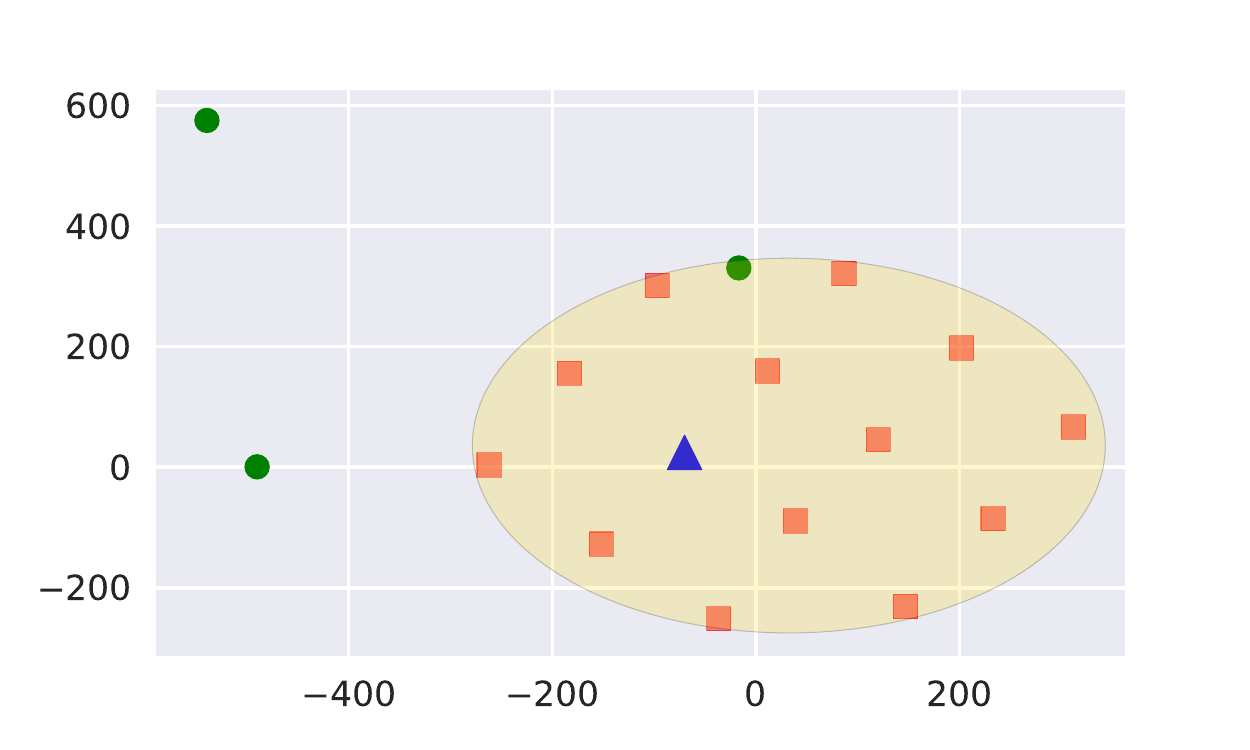} }}%
	\subfloat[\centering 20 extra primitives]{{\includegraphics[scale = 0.4, trim=5 10 20 10, clip]{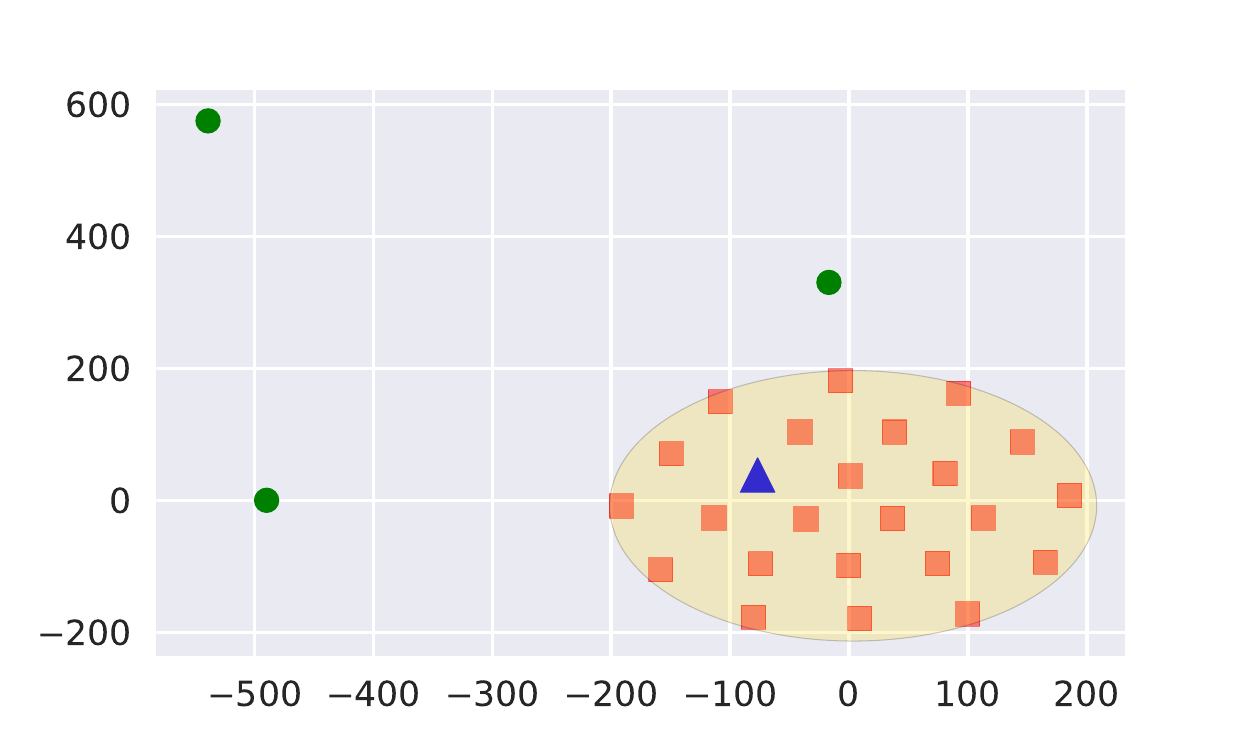} }}%
	\caption{Visualizing the $t$-SNE reduced embeddings of isolated primitive (\markerbluetriangle), example primitives (\markerredsquare) and non-primitives (\markergreencircle) from a learned LSTM model as we increase the number of example primitives in the Colors train set.}%
	\label{fig:colors_lstm_tsne}%
\end{figure*}


Drawing analogy from human vocabulary acquisition \cite{Bloom}, our primitive generalization setting corresponds to the case when a child is explicitly explained the meaning of a word. But children can learn word meaning from implicit usage. In our setting this would translate to using a primitive in a more complex construction, say `jump twice $\to$ JUMP JUMP' instead of the original `jump $\to$ JUMP'. It would be interesting to evaluate how well seq-to-seq models learn the meanings of words from a single sentence and whether they learn to use that word compositionally with other words.  

We consider the `add jump' split in SCAN. Instead of providing the `jump $\rightarrow$ JUMP' primitive definition in the train set, we provide one compositional sentence featuring `jump'. We vary the complexity of this sentence as shown in Table \ref{tab:scan_complexity}. Similar to the case of providing only the primitive definition, we observe that models are unable to generalize and achieve near-zero accuracies. 

We now wish to see whether the presence of more number of primitives and their sentences in the train set helps a model generalize in this scenario (like it did for primitive definitions as shown in Section \ref{sec:prim}). We consider the setup of having 100 primitives and their sentences in the train set (Section \ref{sec:prim}) apart from the one compositional sentence with the word `jump'. We find that models are able to achieve near-perfect generalization accuracies. 

This shows that our idea holds more generally: Adding more primitives and their sentences helps a model effectively learn the meaning of a new primitive, whether specified explicitly via a primitive definition or implicitly in a sentence.

\section{Details of Experimental Setups and Other Results}\label{experiments}

\subsection{Embedding of Isolated Primitive}\label{sec:embedding}

\begin{figure}[t]
	\centering
	\includegraphics[scale=0.48, trim=8 8 10 5, clip]{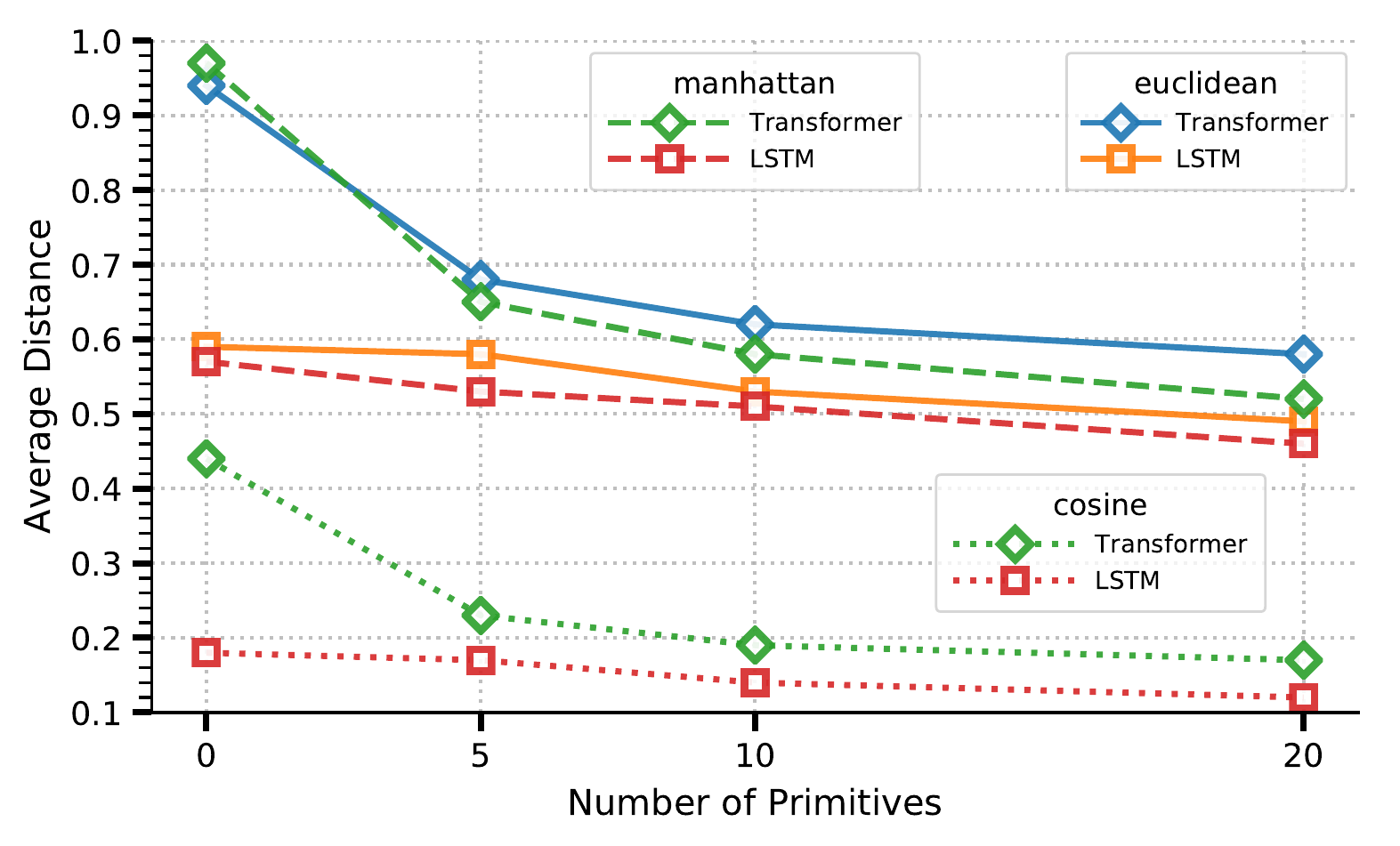}
	\caption{\label{fig:colors_similarity} Measuring the similarity of the embedding of \emph{isolated primitive} with the embeddings of example primitives for learned Transformer and LSTM models as we increase the number of example primitives in the Colors train set.}
\end{figure}

\begin{figure}[t]%
	\centering
	\subfloat[\centering Other Distributions]{{\includegraphics[scale = 0.5, trim=5 10 20 5, clip]{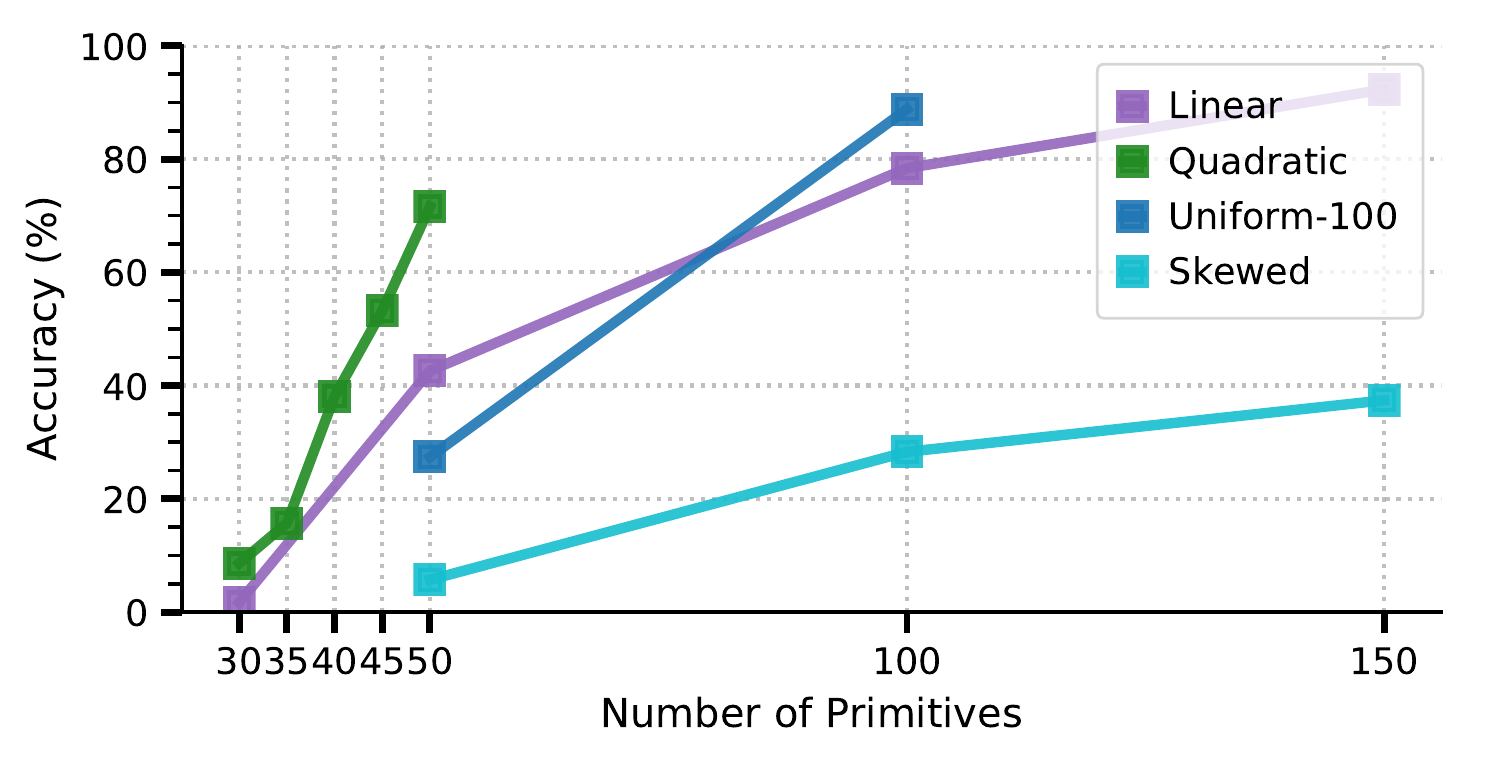} \label{fig:scan_lstm_dist} }}
	
	\subfloat[\centering Uniform Distribution]{{\includegraphics[scale = 0.55, trim=0 0 5 5, clip]{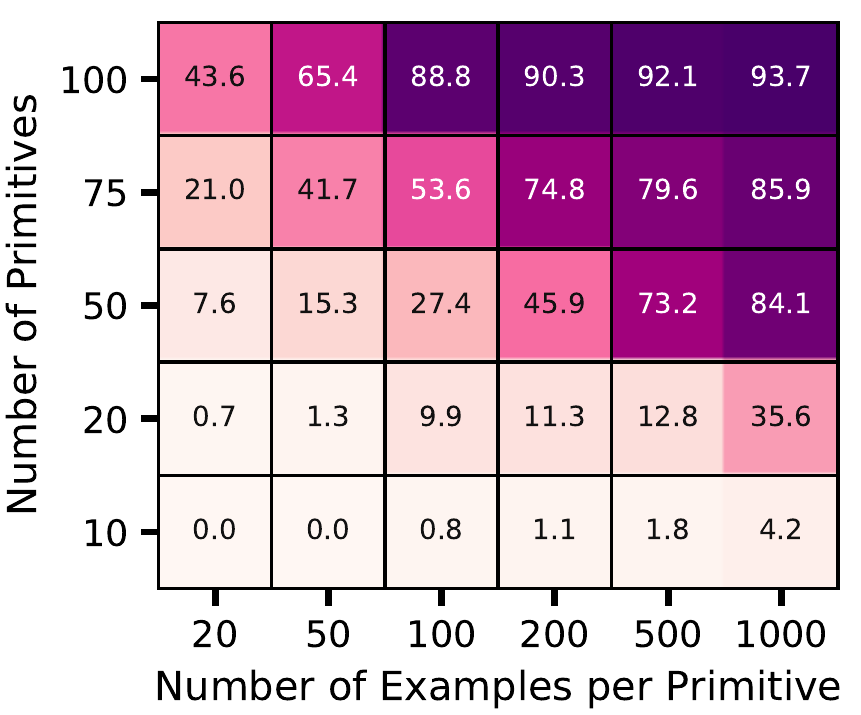} \label{fig:scan_lstm_grid_uni} }}
	
	\caption{Measuring the generalization performance of LSTM on different types of train set distributions of the SCAN dataset.}%
	\label{fig:scan_lstm_distribution}%
\end{figure}

We scale the embedding vectors to unit $L2$-norm for calculating the euclidean distance and unit $L1$-norm for calculating the manhattan distance. For Colors dataset as well, we compare the average distance with other primitives before and after adding primitives to the training data. We again find that as we increase the number of example primitives in the training set, the embedding of the isolated primitive (`zup') gets closer to the example primitives (refer to Fig. \ref{fig:colors_similarity}) in terms of Euclidean, Manhattan and Cosine Distances. 

We additionally show the t-SNE plots of the learned embeddings for the LSTM model on the Colors dataset (Fig. \ref{fig:colors_lstm_tsne}).

\subsection{Impact of Training Distributions}\label{sec:train_dist_appendix}

In Section \ref{sec:train_dist}, we showed results of the Transformer model on various train set distributions of the SCAN dataset. We also experimented with the LSTM model, the results of which can be found in Fig. \ref{fig:scan_lstm_distribution}. We see the same trend as we saw for Transformers.

\subsection{Impact of Model Capacity}\label{sec:capacity_appendix}

\begin{figure}[t]
	\centering
	\includegraphics[scale=0.55, trim=8 8 10 5, clip]{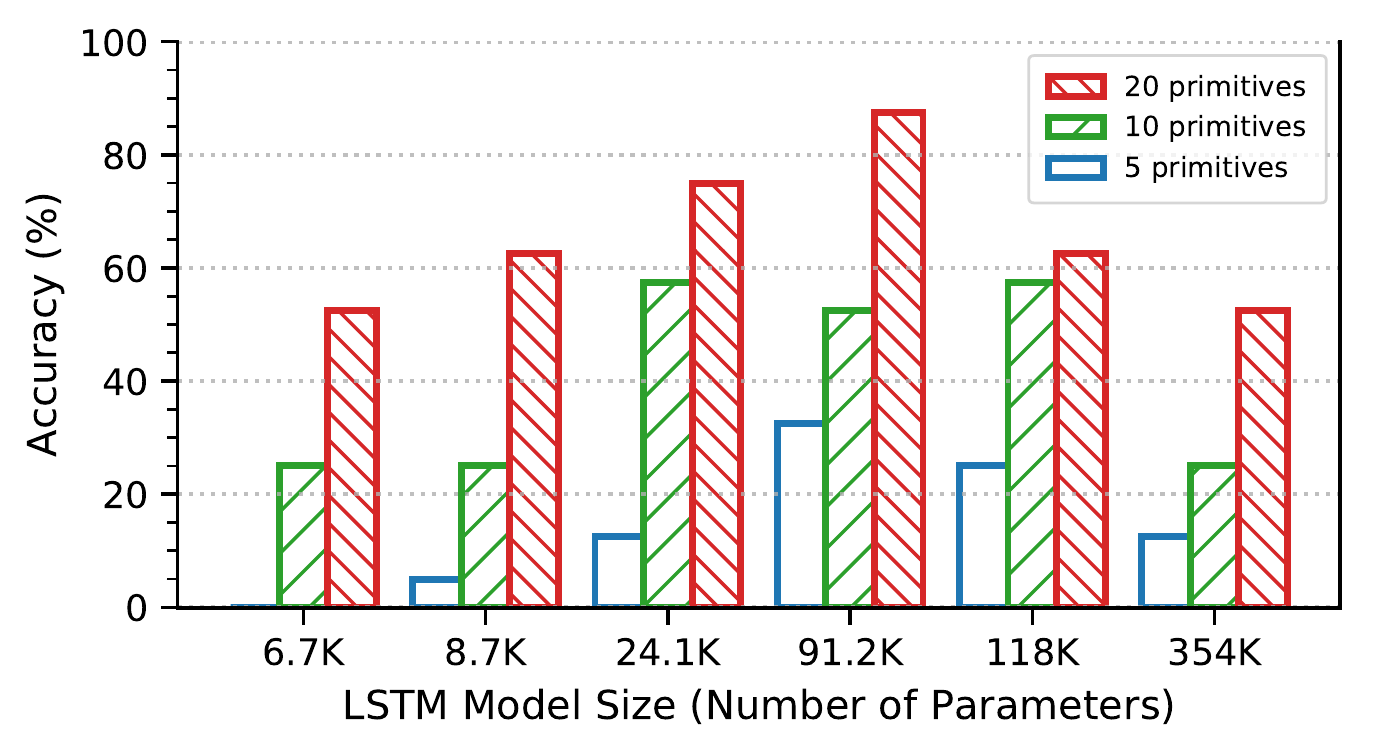}
	\caption{\label{fig:colors_lstm_capacity} Measuring the generalization performance of an LSTM of varying capacity across increasing number of primitives in the Colors train set.}
\end{figure}

In Section \ref{sec:capacity}, we showed results of varying sizes of Transformers trained on datasets with different number of example primitives. We also experimented with the LSTM model, the results of which on the Colors dataset can be found in Fig. \ref{fig:colors_lstm_capacity}. We see the same trend as we saw for Transformers.

\subsection{Variance Across Different Runs}

\begin{figure}[t]
	\centering
	\includegraphics[scale=0.5, trim=8 8 18 15, clip]{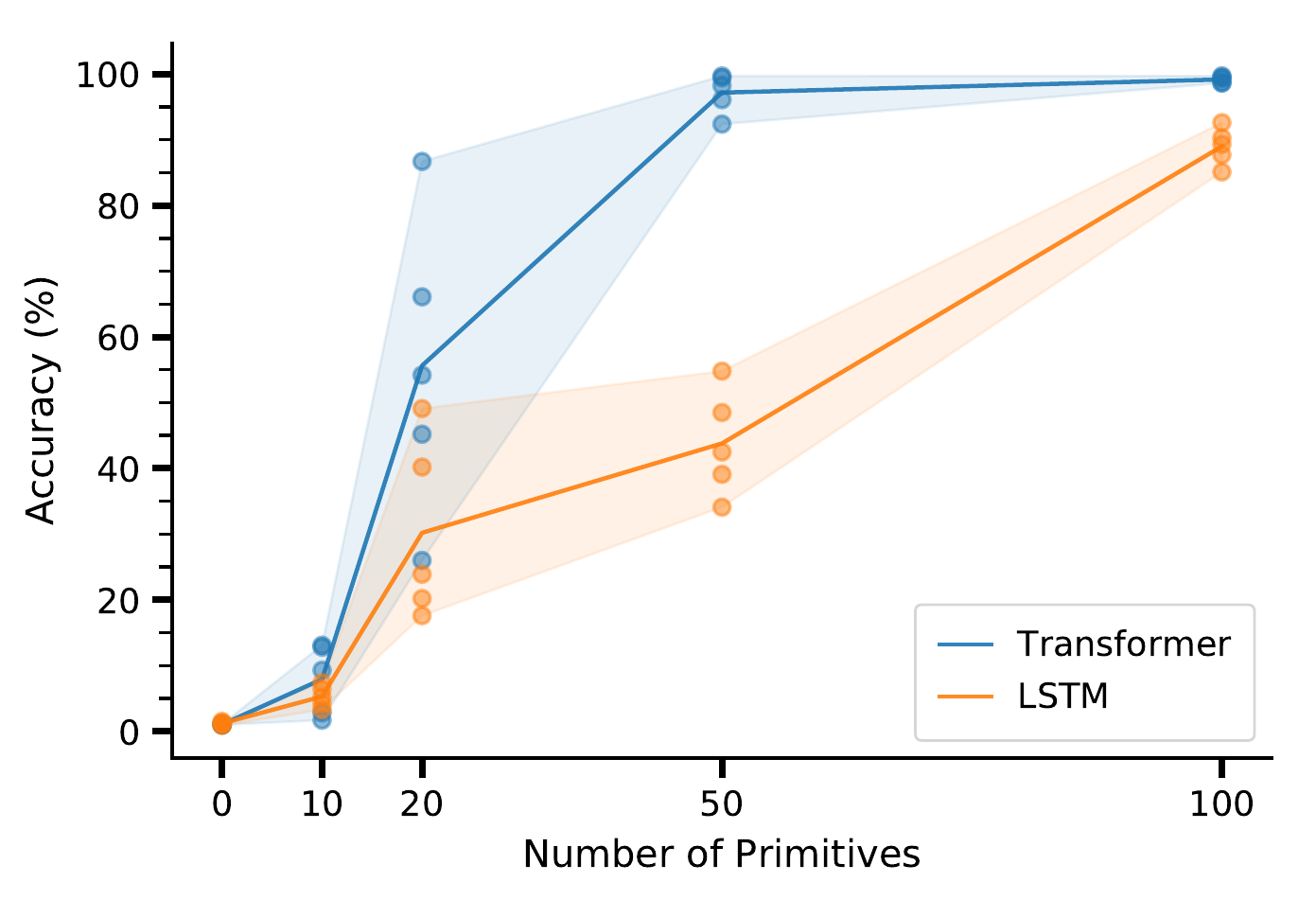}
	\caption{\label{fig:scan_primitives_variance} Generalization performance on SCAN across different runs with random seeds.}
\end{figure}

\begin{figure}[t]
	\centering
	\includegraphics[scale=0.5, trim=8 8 18 15, clip]{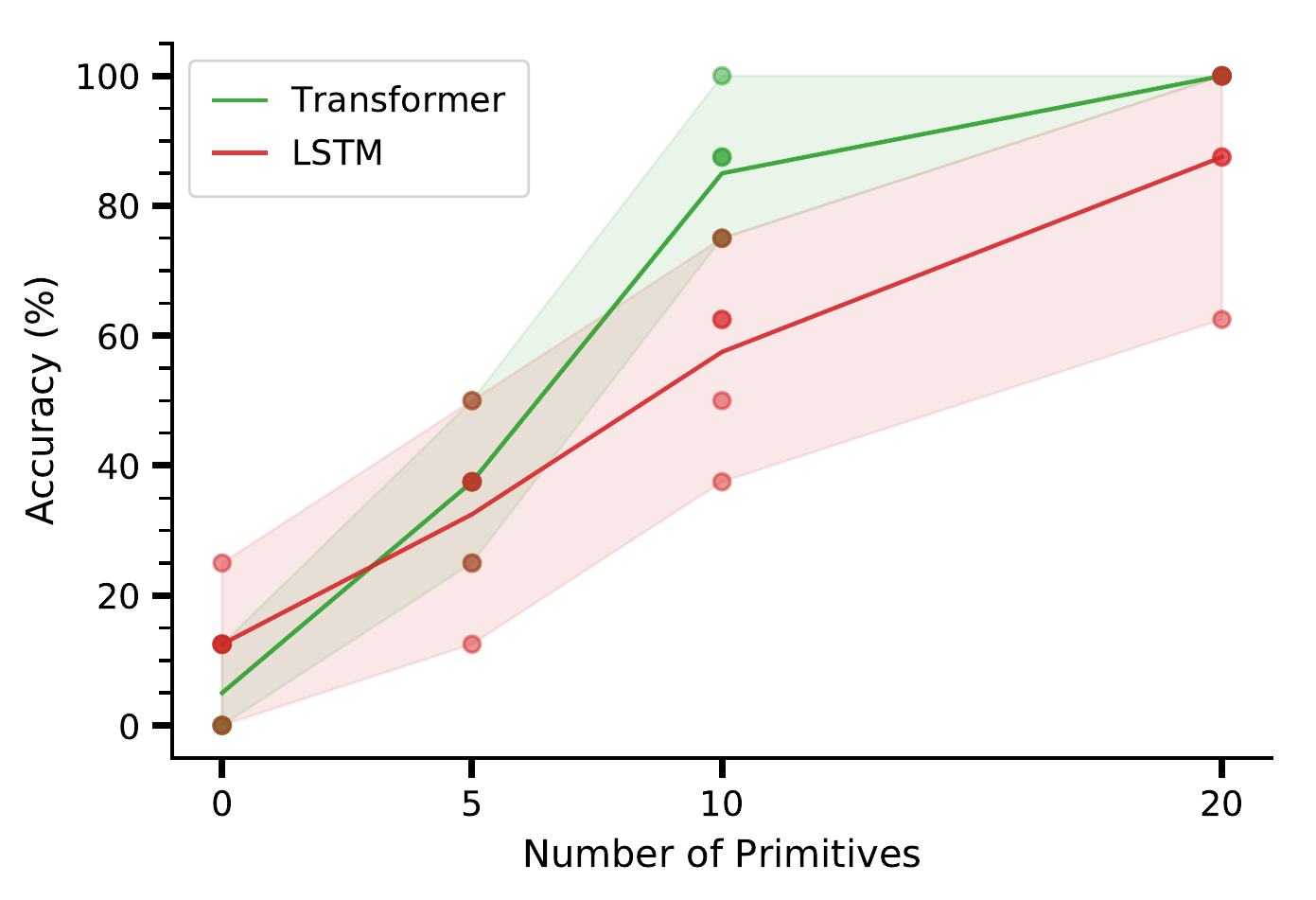}
	\caption{\label{fig:colors_primitives_variance} Generalization performance on Colors across different runs with random seeds.}
\end{figure}

We plot the generalization accuracies of the Transformer and LSTM models on SCAN and Colors datasets over 5 different runs with random seeds in Fig. \ref{fig:scan_primitives_variance}-\ref{fig:colors_primitives_variance}. Both models displayed a high degree of variance in generalization performance on both datasets. It is interesting to see that the variance decreases with increasing number of primitives.

\subsection{Evaluation on Multiple Isolated Primitives}\label{sec:multiple_iso_prims}

Our results are valid not just when there is a single isolated primitive, but even when there are multiple isolated primitives that are used compositionally at test time. While we believe that this holds trivially due to the symmetry of the setup, for completeness, we provide empirical evidence. We consider the setting on SCAN in which the train set has a total of 100 example primitives uniformly distributed. To this train set, in addition to the primitive definition of `jump' (i.e., `jump $\rightarrow$ JUMP'), we add 9 other primitive definitions of newly introduced isolated primitives. Thus, while the size of the train set in this setting was 13185, the size of the new train set is 13194. We then extract templates from the original SCAN test set and exhaustively populate these templates with the 10 isolated primitives. Hence, while the size of the original test set was 7706, the size of the new test set is 77060.

We evaluated Transformers on this data. The best model achieved 94.5\% accuracy on the complete test set, thereby showing that our methodology and results are valid even when there are multiple isolated primitives in the dataset at the same time.

\section{A Note on Other Data Augmentation Methods}\label{sec:sim_work}

Applying data augmentation methods such as GECA \cite{geca} on SCAN will lead to addition of training examples in which the input sentences are compositions of the isolated primitive `jump'. This breaks the systematicity of the setup. While such automatic data augmentation approaches are important resources for enabling compositional generalization, a model that performs well on this modified split cannot be considered to be able to generalize compositionally.

\citet{scan-to-real-data} proposed a data augmentation method based on the theory of meaningful learning. Similar to our work, they also augment the train set by adding more primitives (e.g. `jump\_0', `jump\_1', ..., `jump\_n'). However, compared to our work, their setup is completely different: The new primitives that they add to the train set are all still mapped to the output token of an example primitive `jump', which is `JUMP' (i.e. `jump\_0 $\rightarrow$ JUMP', ..., `jump\_n $\rightarrow$ JUMP'). Their train set has examples showing compositions of `jump' while their test set evaluates for novel compositions of the newly added primitives. We argue that their setup cannot be considered one-shot primitive generalization since now the model can see the output token `JUMP' in composition with other words. We claim that this familiarity with the output token enables a model to generalize well on the test data even if the newly added primitives are only presented one-shot in the train set. Indeed, \citet{scan} also suggested that the reason why models are able to do well on the `turn left' split of SCAN is because the train set consists of many examples that have the output token `LTURN' used compositionally. 

To validate our claim, we propose a simple experiment. In the original SCAN `add jump' split, we map `jump $\rightarrow$ WALK' instead of `jump $\rightarrow$ JUMP' for all examples (primitive definition as well as compositional sentences) in both the train and test sets. In this setup, even though the input word `jump' is seen only once at train time, it's mapping `WALK' is used compositionally in many examples. On evaluating a Transformer model on this split, we found that it achieves a near-perfect accuracy. This shows that providing compositional examples with the output token of the isolated primitive not only breaks systematicity, but is the reason behind the high performance of models in that setting.

\end{document}